\pdfoutput=1

\documentclass[11pt]{article}

\usepackage[]{acl}
\usepackage{times}
\usepackage{latexsym}
\usepackage{url}

\usepackage{xspace}
\usepackage{graphicx}
\usepackage{multirow}
\usepackage{makecell}
\usepackage{lscape}
\usepackage{bm}
\usepackage{algorithm}
\usepackage{algpseudocode}
\usepackage{framed}
\usepackage{color}
\usepackage{xcolor}
\usepackage{booktabs,colortbl}
\usepackage{bbm}
\usepackage{subfigure}
\usepackage{multirow} 
\usepackage{amssymb}
\usepackage{amsmath}
\usepackage{soul}
\usepackage{multirow}
\usepackage{tabularx}

\usepackage{booktabs}
\usepackage{setspace}
\usepackage{utfsym}

\newcommand{\eg}{\emph{e.g.,}\xspace}

\newcommand{\etc}{\emph{etc.}\xspace}

\newcommand{\baby}{C$^3$\textsc{da}\xspace}

\usepackage[T1]{fontenc}

\usepackage[utf8]{inputenc}

\usepackage{microtype}

\title{A Contrastive Cross-Channel Data Augmentation Framework for Aspect-Based Sentiment Analysis}

\author{Bing Wang$^{1}$\thanks{~~Work was done when Bing and Qihuang were interning at JD Explore Academy.}~, \ Liang Ding$^{2}$\thanks{~~Liang Ding is the corresponding author.}, \ Qihuang Zhong$^{4}$, \ Ximing Li$^{1}$, \ Dacheng Tao$^{2,3}$\\
$^{1}$College of Computer Science and Technology, Jilin University\\ \ $^{2}$JD Explore Academy \ $^{3}$The University of Sydney \ $^{4}$Wuhan University\\
\texttt{dingliang1@jd.com}, \texttt{liximing86@gmail.com}
}
\begin{document}
\maketitle
\begin{abstract}
Aspect-based sentiment analysis (ABSA) is a fine-grained sentiment analysis task, which focuses on detecting the sentiment polarity towards the aspect in a sentence. However, it is always sensitive to the \textit{multi-aspect challenge}, where features of multiple aspects in a sentence will affect each other.
To mitigate this issue, we design a novel training framework, called Contrastive Cross-Channel Data Augmentation (\baby), 
which leverages an in-domain generator to construct more multi-aspect samples and then boosts the robustness of ABSA models via contrastive learning on these generated data. 
In practice, given a generative pretrained language model and some limited ABSA labeled data, we first employ some parameter-efficient approaches to perform the in-domain fine-tuning. Then, the obtained in-domain generator is used to generate the synthetic sentences from two channels, i.e., Aspect Augmentation Channel and Polarity Augmentation Channel, which generate the sentence condition on a given aspect and polarity respectively. Specifically, our \baby performs the sentence generation in a cross-channel manner to obtain more sentences, and proposes an Entropy-Minimization Filter to filter low-quality generated samples.
Extensive experiments show that our \baby can outperform those baselines without any augmentations by about 1\% on accuracy and Macro-F1. {Code and data are released in \url{https://github.com/wangbing1416/C3DA}.}

\end{abstract}

\section{Introduction}

Sentiment Analysis (SA) is a typical Natural Language Understanding (NLU) task to predict the sentence-level sentiment polarities \cite{pang2007opinion,liu2012sentiment}. However, since a single review sentence always exists multiple polarities, \textbf{A}spect-based \textbf{S}entiment \textbf{A}nalysis (\textbf{ABSA}), a fine-grained sentiment analysis task \cite{ma2017interactive,fan2018multi,sun2019aspect,wang2020relational,li2021dual,zhang2022a}, is further introduced to detect the sentiment polarities towards given aspects (entities) in a review sentence. Gathering a review "while the \underline{ambiance} and \underline{atmosphere} were great, the \underline{food} and \underline{service} could have been a lot better.", the sentiment polarities of \textit{"ambiance"} and \textit{"atmosphere"} are \textit{Positive} and we get \textit{Negative} polarity for aspects \textit{"food"} and \textit{"service"}.

It is crucial to capture the aspect-specific contextual features for an ABSA model. Unfortunately, as shown in the aforementioned example, the multi-aspect challenge that there are multiple aspects in a sentence always deteriorates the model's generalization, especially when the expressed polarities of multiple aspect words in a sentence are opposite. 
Meanwhile, since ABSA is a low-resource task, it is hard to train a robust model under fewer samples.

Some existing works focused on the multi-aspect challenge,
such as \citet{lu2011multi,hu2019can} in which novel model structures were designed to capture aspect-specific sentiment information. Additionally, \citet{jiang2019a} manually collected a large-scale high-quality multi-aspect multi-sentiment (MAMS) dataset.
Undoubtedly, the data-centric MAMS is an effective approach to tackle the multi-aspect problem, but the human-annotated and non-expandable limitations still hinder the robust training for ABSA models.
Therefore, given limited labeled data, it is critical to investigate how to collect more in-domain multi-aspect samples automatically. On the other hand, capturing the aspect-specific information from multi-aspect sentences is also necessary for robust ABSA models.


In response to the aforementioned problems, we propose a novel training framework, namely \textbf{C}ontrastive \textbf{C}ross-\textbf{C}hannel \textbf{D}ata \textbf{A}ugmentation (\textbf{\baby}), to generate more in-domain multi-aspect samples and train robust ABSA models based on these generated data. Firstly, inspired by successful generative pretraining models~\cite{raffel2020exploring, lewis2020bart}, we employ an representative pretraining models T5~\cite{raffel2020exploring} as the multi-aspect data generator. In practice, due to the limited ABSA labeled data, it is hard to fine-tuning the entire T5 model effectively. Thus, we further introduce some parameter-efficient methods, e.g. prompt~\cite{lester2021power}, prefix~\cite{li2021prefix} and LoRA~\cite{hu2021lora}, to tune the T5 generator. In this way, a domain -specific generator is obtained and we can apply it to collect more in-domain multi-aspect samples.


To be more specific, there are two channels to generate expected multi-aspect samples in our framework, i.e. \textbf{A}spect \textbf{A}ugmentation \textbf{C}hannel (\textbf{AAC}) and \textbf{P}olarity \textbf{A}ugmentation \textbf{C}hannel (\textbf{PAC}). In the AAC, the generator is encouraged to generate the synthetic sentence towards the given sentence and aspect, while in the PAC, the synthetic sentence towards the given sentence and polarity is obtained.
To further boost the generated samples, we attempt to collect the multi-aspect samples in a cross-channel manner. Specifically, 
in the first generation, the source sentence is fed to the double channels and obtain an aspect-specific sentence and a polarity-inverted sentence respectively. In the second generation, both generated sentences are injected to the another channel to get the ultimate sentences. 
Finally, we propose an Entropy-Minimization Filter (EMF) to filter some low-quality generated samples. And a contrastive training objective that can draw away the different aspect's embeddings in a sentence is also leveraged to alleviate the multi-aspect problem and help train the final robust ABSA model.

We conduct sufficient experiments on three popular ABSA datasets, i.e. \textit{Restaurant}, \textit{Laptop} \cite{pontiki2012semeval} and \textit{Twitter} \cite{dong2014adaptive}, to prove that our \baby framework can boost the model's robustness and predictive performance, and outperform other NLP data augmentation strategies. Specifically, with the help of our \textbf{\baby}, RoBERTa-based models can achieve averaged performance improvement 1.28\% and 1.62\% in terms of accuracy and macro-F1 respectively, while the improvements of BERT-based models are also over 0.87\% and 1.10\%. Some in-depth discussions and case studies are also executed. Contributions of this paper are threefold:
\begin{itemize}
    \item We recast the vanilla ABSA training scheme with a data augmentation-based training framework and a contrastive training objective to tackle the multi-aspect challenge.
    \item We design a novel cross-channel data augmentation method based on generative large-scale pretrained language models to generate high-quality in-domain multi-aspect samples, which has great potential to benefit other fine-grained NLU tasks.
    \item Extensive experiments on three widely-used datasets show the effectiveness of our \baby.
\end{itemize}

\section{Related Work}

\begin{figure*}[htbp]
  \centering
  \includegraphics[scale=0.54]{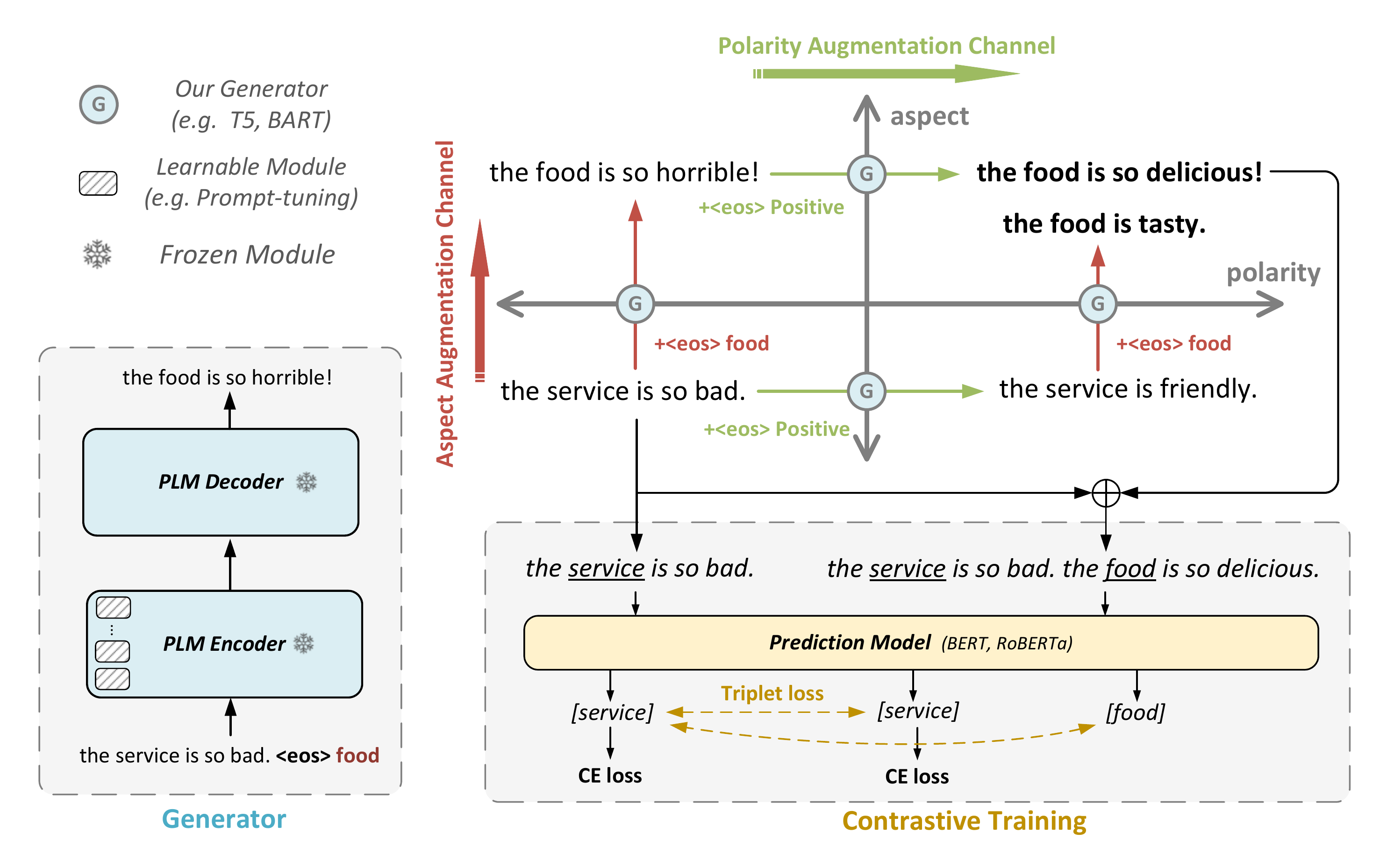}
  \caption{The overall \baby framework. In our generator, learnable modules can be optimized in the fine-tuning stage and will also be frozen in the generation stage, and the PLM encoder-decoder structure is always fixed.}
  \label{framework}
\end{figure*}

\subsection{Language Models and Parameter Efficient}

Large Pre-trained Language Models (PLMs) are still research cores in various natural language processing tasks. The PLMs tend to be variants of Transformer \cite{vaswani2017attention}, and are trained on massive unlabeled raw sentences under some linguistic unsupervised objectives.
According to the model structure, these Transformer-based PLMs are roughly divided into (1) Encoder-only LMs  \cite{devlin2019bert,liu2019roberta} can be utilized to capture token-level and sentence-level features, (2) Decoder-only LMs \cite{radford2018improving,radford2019language} tend to design an auto-regressive objective to cater to text generation tasks and (3) Encoder-Decoder LMs \cite{lewis2020bart,raffel2020exploring} need two sentences (source and target) to perform conditional generation.

The above PLMs always follow the \textit{pre-train + fine-tune (FT)} paradigm to adapt the downstream tasks, however, tuning a whole model is a tricky challenge and is not applicable to low-resource NLP tasks. Therefore, a novel \textit{pre-train + prompt-tune (PT)} \cite{liu2021pre} concept is proposed to break this challenge, PT-based models can also be divided into (1) manual discrete prompts design human-made templates to adapt different pre-trained objective \cite{liu2021solving}, and (2) soft prompts inject some learnable pseudo tokens or matrices to frozen PLMs \cite{liu2021gpt,lester2021the,li2021prefix,zhong2022panda}.

\subsection{Aspect-based Sentiment Analysis}

Aspect-based Sentiment Analysis (ABSA) is a kind of Text Classification task, that benefits from a better aspect-aware text representation. Therefore, an amount of neural network-based models were proposed to break the ABSA challenge, for example, graph-based models \cite{sun2019aspect,wang2020relational,li2021dual,zhong2022knowledge} conduct graph convolutional operations on dependency trees to encode the semantic information, and attention-based models \cite{ma2017interactive,chen2017recurrent,fan2018multi,song2019attentional,liu2021unified} focus on interaction between aspect terms and context tokens, \etc

In recent years, pre-trained language models changed the situation of ABSA because they are better text encoders and can capture context features. Therefore, researchers explored more views to improve ABSA performance, except that PLM is regarded as an encoder to replace the vanilla \textit{Glove + BiLSTM} paradigm. For example, \newcite{dai2021does} leverages RoBERTa to re-construct dependency trees, \newcite{yan2021a,li2021sentiprompt,zhang2021towards} try to integrate various ABSA subtasks into a unified generative framework, and \newcite{zhou2021implicit,seoh2021open} combine ABSA with other NLP tasks, \etc

\subsection{Data Augmentation}

To improve the scale of training samples, some Data Augmentation (DA) methods in the NLP community are proposed~\cite{mixup,wei2019eda,kobayashi_contextual,wu2019conditional,Anaby2020do,ding-etal-2021-improving,Cao2021TowardsED,wang2022promda,zhang2022bliss}. EDA \cite{wei2019eda} transforms source sentences with Synonym Replacement approaches, \etc CBERT \cite{wu2019conditional} introduces a new conditional masked language model task. Meanwhile, some PLM-based generative augmentation frameworks \cite{Anaby2020do,wang2022promda,zhang2022bliss} also have been explored. Also, recent work reveals the complementarity between PLMs and classical approaches, e.g., back translation~\cite{liu-etal-2021-complementarity-pre} and random initialization~\cite{zan2022PTvsRI}, to augment both the accuracy and generalization.

\section{Our \baby Framework}

\paragraph{Task Description of ABSA}
Given a group of triplets $\Omega^{src} = \{\mathbf{s}_i, \mathbf{a}_i, \mathbf{p}_i\}_{i=1}^N$, a sentence $\mathbf{s}_i = \{s_{i1}, \cdots, s_{iL}\}$ and an aspect-span indicator $\mathbf{a}_i \in \{0,1\}^L$ ($1$ denotes this token is an aspect, vice versa) will be fed to a trainable model to obtain a hidden embedding $\mathbf{h}_i = \mathcal{F}(\mathbf{s}_i, \mathbf{a}_i, \Theta) \in \mathbb{R}^{L \times D}$, where $N$, $L$ and $D$ denote the instance scale, sentence length and hidden embedding size, and $\mathcal{F}(\cdot, \Theta)$ is a prediction model (\eg BERT \cite{devlin2019bert}). Then, a ground-truth polarity label $\mathbf{p}_i \in \{0, 1, 2\}$ is utilized to conduct supervised learning to estimate task-specific parameters $\Theta$.

\begin{table*}[ht]
\centering
\renewcommand\arraystretch{1.2}
\small
  \caption{An example of the concatenation operation.}
  \label{examplecat}
  \setlength{\tabcolsep}{5pt}{
  \begin{tabular}{m{2.3cm}<{\centering}m{11.3cm}<{\centering}}
    \toprule
    \rowcolor{lightgray} $\mathbf{s}$ & \textit{I've been here 3 times for lunch and it is one of my favorites in the city.} \\
    $[\mathbf{s}, \textit{service}]$ & \textit{I've been here 3 times for lunch and it is one of my favorites in the city.} $<eos>$ \textit{\textbf{service}} \\
    $[\mathbf{s}, \textit{Negative}]$ & \textit{I've been here 3 times for lunch and it is one of my favorites in the city.} $<eos>$ \textit{\textbf{so bad}} \\
    \bottomrule
  \end{tabular} }
\end{table*}

\subsection{Overall Learning Objective}
In this section, we introduce the overall process of our proposed method, and more details will be further described in Section~\ref{sec3.2} and \ref{sec3.3}, and our \baby framework is shown in Figure~\ref{framework}.

Our \baby framework consists of two steps: (1) Data Augmentation and (2) Contrastive Training. We utilize the representative pre-trained T5 model \cite{raffel2020exploring} as our generator in the data augmentation stage\footnote{Note that we can employ various generative pretraining models as the generator. T5 is used in this work, since it is widely-used and powerful.}. Given a sentence $\mathbf{s}_i$, we expect to generate an another sentence $\mathbf{\widehat s}_i = \mathcal{G}(\mathbf{s}_i; \Pi)$ that expresses the opposite polarity and has a different aspect from the source sentence, where $\mathcal{G}(\cdot; \Pi)$ is the T5 generator.

Basically, we conduct the Supervised Classification Training (SCT) on source sentences and augmented sentences, the objective is as follows: 

\begin{align}
\label{Eq1}
    \mathcal{L}_{SCT} = \frac{1}{N} \sum \nolimits _{i=1}^N & \ell_{CE}(\mathbf{h}_i \mathbf{W}_s + \mathbf{b}_s, \mathbf{p}_i) \nonumber \\
    + \alpha & \ell_{CE}(\mathbf{h}_i^p \mathbf{W}_s + \mathbf{b}_s, \mathbf{p}_i),
\end{align}

\begin{equation}
\label{Eq2}
    \mathbf{h}_i^p = \mathcal{F} \left( [\mathbf{s}_i, \mathbf{\widehat s}_i], \mathbf{a}_i, \Theta \right),
\end{equation}
where $[\cdot, \cdot]$ and $\ell_{CE}(\cdot)$ are the concatenate operation and a Cross-Entropy loss function, $\mathbf{W}_s \in \mathbb{R}^{D \times M}$ and $\alpha$ denote the classification head and a hyper-parameter, respectively.

To obtain more robust performance, we leverage the Contrastive Training (CT) formulated as the triplet loss \cite{tripletloss2003} as follows:

\begin{align}
\label{Eq3}
    \mathcal{R}_{CT} = \frac{1}{N} \sum \nolimits _{i=1}^N max\{ & d(\mathbf{h}_i, \mathbf{h}_i^p) - \nonumber \\
    & d(\mathbf{h}_i, \mathbf{h}_i^n) + \xi, 0\},
\end{align}
where $\mathbf{h}_i^n$ is the average-pooled sentence representation of $\mathbf{\widehat s}_i$ in the sentence $[\mathbf{s}_i,  \mathbf{\widehat s}_i]$, $\xi$ is a hyper-parameter called margin that can control whether an instance should be trained, and $d(\cdot, \cdot)$ denotes a distance measure function. Specially, we apply a negative cosine similarity to measure the embedding distance $d(\cdot, \cdot)$.
\begin{equation}
\label{Eq7}
    d(\mathbf{h}_i, \mathbf{h}_i^p) = - \frac{\mathbf{h}_i \mathbf{h}_i^p}{\|\mathbf{h}_i\| \times \|\mathbf{h}_i^p\|}.
\end{equation}

Finally, the overall learning objective can be formulated as:
\begin{equation}
\label{Eq4}
    \mathcal{L} = \mathcal{L}_{SCT} + \beta \mathcal{R}_{CT},
\end{equation}
where $\beta$ is a controllable hyper-parameter.

\subsection{Cross-Channel Data Augmentation} \label{sec3.2}
Our motivation is to generate high-quality in-domain multi-aspect samples.
In practice, the cross-channel data augmentation consists of two steps: 1) tune the pretrained backbone model to adapt our cross-channel sentence generation; 2) generate sentences with opposite polarity and different aspects.

\paragraph{Pre-Trained Generator}
We employ pre-trained encoder-decoder model T5 as our backbone. 
ABSA is a low-resource task, therefore, tuning a whole large language model (\eg T5) with scarce samples will cause an over-fitting problem. 
Inspired by \citet{lester2021power,wang2022promda}, we utilize the parameter-efficient strategies to tune the generation model. 
In particular, we consider several efficient tuning methods, including prefix-tuning~\cite{li2021prefix}, prompt-tuning~\cite{lester2021power}, and LoRA~\cite{hu2021lora}. We grid-search the optimal efficient tuning method for each dataset, detailed comparisons can be found at Appendix~\ref{AppendixA}.

\paragraph{Model Fine-Tuning}
To make the T5 model capable of generating new sentences conditioned on the original sentence and given aspect or polarity, we first construct the fine-tuning data based on the ABSA training set.
Concretely, we randomly sample two instances $\{\mathbf{s}_i\}$ and $\{\mathbf{s}_j, \mathbf{a}_j, \mathbf{p}_j\}$ from the source triplets $\Omega^{src}$. Then, we can tune the T5 model to generate $\mathbf{s}_j$ conditioned on  $[\mathbf{s}_i,\mathbf{a}_j]$ or $[\mathbf{s}_i, \mathbf{p}_j]$.
As aforementioned, parameter-efficient strategies are used to avoid over-fitting. Taking the cutting-edge prompt-tuning~\cite{lester2021power} as an example, the massive parameters in the generator should be fixed in this stage, and we only tune the parameters of prompts.

\paragraph{Sentence Generation}
The sentence generation for data augmentation stage consists of \textbf{A}spect \textbf{A}ugmentation \textbf{C}hannel (\textbf{AAC}) and \textbf{P}olarity \textbf{A}ugmentation \textbf{C}hannel (\textbf{PAC}).

AAC generates an aspect-specific sentence $\mathbf{s}_i^a = \mathcal{G}([\mathbf{s}_i, \mathbf{\widehat A}], \Pi)$\footnote{AAC can generate one or more sentences, we take generating a single sentence as an example.}, where $\Pi$ is the trainable parameters of the T5 generator $\mathcal{G}(\cdot)$. PAC generates a polarity-inverted sentence $\mathbf{s}_i^p = \mathcal{G}([\mathbf{s}_i, \mathbf{\widehat P}], \Pi)$, where $\mathbf{\widehat A}$ is a random-sampled in-domain aspect and $\mathbf{\widehat P}$ is a group of seed spans that have the inverted polarity from $\mathbf{s}_i$, such as \textit{so good}. Then, these sentences will be fed into the another channel to obtain $\mathbf{s}_i^{pa} = \mathcal{G}([\mathbf{s}_i^p, \mathbf{\widehat A}], \Pi)$ and $\mathbf{s}_i^{ap} = \mathcal{G}([\mathbf{s}_i^a, \mathbf{\widehat P}], \Pi)$ that can satisfy our motivation. We give an example of the concatenation operation $[\mathbf{s},\cdot]$ in Table~\ref{examplecat}.

Note that for sentence generation, both the generator backbone and prompts will be frozen. 
The detailed Cross-Channel Data Augmentation method is shown in Algorithm~\ref{alg}.

\begin{algorithm}[htb]
  \caption{Cross-Channel Data Augmentation.}
  \setstretch{1.3}
  \small
  \label{alg}
  \begin{algorithmic}[1]
    \Require
      Source triplets $\Omega^{src} = \{\mathbf{s}_i, \mathbf{a}_i, \mathbf{p}_i\}_{i=1}^N$;
      Initial pre-trained T5 generator with soft prompt $\mathcal{G}(\cdot, \Pi^0)$;
      Pre-defined aspect set $\mathcal{A}$;
      Fine-tune iterations $\tau$.
    \Ensure
      Augmented sentences $\Omega^{aug} = \{\mathbf{\widehat s}_i\}_{i=1}^N$.
    \For{$iter$ in $\tau$} \algorithmiccomment{Stage \uppercase\expandafter{\romannumeral1}: Model Fine-tuning}
    \State $\{\mathbf{s}_i\}, \{\mathbf{s}_j, \mathbf{a}_j, \mathbf{p}_j\} \gets Sample(\Omega^{src}) $
    \State $\nabla^a \gets loss\left(\mathbf{s}_j, \mathcal{G}([\mathbf{s}_i, \mathbf{a}_j], \Pi^{iter})\right)$
    \State $\nabla^p \gets loss\left(\mathbf{s}_j, \mathcal{G}(\left[\mathbf{s}_i, \mathbf{p}_j \right], \Pi^{iter})\right)$
    \State  $\Pi^{iter + 1} \gets Train(\Pi^{iter}, \nabla^a, \nabla^p)$
    \EndFor \\
    $\Pi \gets \Pi^{\tau}$
    
    \For{$\{\mathbf{s}_i, \mathbf{a}_i, \mathbf{p}_i\}$ in $\Omega^{src}$}
    \State \algorithmiccomment{Stage \uppercase\expandafter{\romannumeral2}: Sentence Generation}
    \State $\mathbf{\widehat A} \gets Sample(\mathcal{A})$
    \State $\mathbf{\widehat P} \gets - \mathbf{p}_i$ \algorithmiccomment{$- \mathbf{p}$ denotes opposite polarity from $\mathbf{p}$}
    \State $\mathbf{s}_i^a \gets \mathcal{G} \left( [\mathbf{s}_i, \mathbf{\widehat A}], \Pi \right)$, $\mathbf{s}_i^p \gets \mathcal{G} \left( [\mathbf{s}_i, \mathbf{\widehat P}], \Pi \right)$ 
    \State $\mathbf{s}_i^{pa} \gets \mathcal{G} \left( [\mathbf{s}_i^p, \mathbf{\widehat A}], \Pi \right)$, $\mathbf{s}_i^{ap} \gets \mathcal{G} \left( [\mathbf{s}_i^a, \mathbf{\widehat P}], \Pi \right)$ 
    \State $\Omega^{aug} \gets \Omega^{aug} + Filter(\mathbf{s}_i^a, \mathbf{s}_i^p, \mathbf{s}_i^{pa}, \mathbf{s}_i^{ap})$
    \EndFor \\
    \Return $\Omega^{aug}$;
  \end{algorithmic}
\end{algorithm}


\subsection{Entropy-Minimization Filtering} \label{sec3.3}

Cross-Channel Data Augmentation generates four sentences for each source sentence, and we can leverage these sentences to train a prediction model $\mathcal{F}(\cdot, \Theta)$. 
However, it is necessary to filter out some low-quality samples that carry noisy (or difficult) information, which can be estimated by language modeling~\cite{moore-lewis-2010-intelligent} and norm of word embedding~\cite{liu-etal-2020-norm}. 

To ensure the certainty of synthetic sentences, we design an Entropy-Minimization Filter (EMF) to filter those noisy sentences.
We establish a hypothesis that \textbf{\textit{noisy sentences always express an ambiguous polarity}} such that our prediction model gives a more smooth polarity probability distribution. 
To be specific, the information entropy of prediction distribution as Equation~\ref{Eq5} is an appropriate measure. 
A sentence with larger prediction entropy should be filtered, and a hyper-parameter $k$ can control how many augmented sentences are selected to participate in training for each sample.
\begin{equation}
\label{Eq5}
    \mathcal{H}_i = - \mathbb{E}_{\mathbf{y}_i} \big[ \log_2 p(\mathbf{y}_i) \big] = - p(\mathbf{y}_i) \log_2 p(\mathbf{y}_i),
\end{equation}

\begin{equation}
\label{Eq6}
    p(\mathbf{y}_i) = softmax (\mathbf{h}_i \mathbf{W}_s + \mathbf{b}_s).
\end{equation}

\section{Experiment}

\begin{table*}[t]
\centering
\renewcommand\arraystretch{1.2}
\small
  \caption{Empirical results of \textbf{\baby} and other cutting-edge data augmentation methods. The best scores of each metric are indicated in bold, and $\uparrow$ denotes the improvement over BERT-\textit{base} and RoBERTa-\textit{base} baselines.}
  \label{result}
  \setlength{\tabcolsep}{5pt}{
  \begin{tabular}{l m{1.3cm}<{\centering}m{1.4cm}<{\centering}m{1.4cm}<{\centering}m{1.4cm}<{\centering}m{1.4cm}<{\centering}m{1.4cm}<{\centering}}
    \toprule
    \multirow{2}{*}{Model} & \multicolumn{2}{c}{\textit{Restaurant}} &  \multicolumn{2}{c}{\textit{Laptop}} & \multicolumn{2}{c}{\textit{Twitter}} \\
    \cmidrule(lr){2-3}\cmidrule(lr){4-5}\cmidrule(lr){6-7}
    & acc & F1 & acc & F1 & acc & F1 \\
    \midrule
    BERT-\textit{base} & 86.31 & 80.22 & 79.66 & 76.11 & 76.50 & 75.23 \\
    + EDA~\cite{wei2019eda} & 86.42 & 79.63 & 79.59 & 75.79 & 76.26 & 75.16 \\
    + BT~\cite{fan2021beyond} & 86.47 & 80.29 & 79.67 & 76.35 & 76.63 & 75.47 \\
    + CBERT~\cite{wu2019conditional} & 86.27 & 80.00 & 79.83 & 76.12 & 76.44 & 75.36 \\
    + SCon~\cite{liang2021enhancing} & 86.51 & 80.55 & 80.23 & 76.48 & - & -\\
    \rowcolor{lightgray} + \textbf{\baby} (Ours) & \textbf{86.93}$_{\uparrow0.62}$ & \textbf{81.23}$_{\uparrow1.01}$ & \textbf{80.61}$_{\uparrow0.95}$ & \textbf{77.11}$_{\uparrow1.00}$ & \textbf{77.55}$_{\uparrow1.05}$ & \textbf{76.53}$_{\uparrow1.30}$ \\
    \midrule
    RoBERTa-\textit{base} & 86.38 & 80.29 & 80.10 & 76.24 & 76.63 & 75.37 \\
    + EDA~\cite{wei2019eda} & 86.43 & 80.21 & 80.38 & 76.59 & 76.47 & 75.36 \\
    + BT~\cite{fan2021beyond} & 86.50 & 80.59 & 80.22 & 76.73 & 76.59 & 75.47 \\
    + CBERT~\cite{wu2019conditional} & 86.77 & 80.51 & 80.54 & 76.57 & 76.73 & 75.37 \\
    \rowcolor{lightgray} + \textbf{\baby} (Ours) & \textbf{87.11}$_{\uparrow0.73}$ & \textbf{81.63}$_{\uparrow1.34}$ & \textbf{81.83}$_{\uparrow1.73}$ & \textbf{78.46}$_{\uparrow2.22}$ & \textbf{78.31}$_{\uparrow1.38}$ & \textbf{76.67}$_{\uparrow1.30}$ \\
    \bottomrule
  \end{tabular} }
\end{table*}

\subsection{Experimental Settings and Baselines}


We conduct our experiments on three public datasets, i.e. \textit{Restaurant}, \textit{Laptop} from SemEval 2014 ABSA task ~\cite{pontiki2012semeval} and \textit{Twitter} ~\cite{dong2014adaptive}.
In practice, we run the experiments over 5 random seeds and report the average values of accuracy (acc) and Macro-F1 (F1) to avoid stochasticity.

Two widely-used pretraining models, i.e. BERT ~\cite{devlin2019bert} and RoBERTa ~\cite{liu2019roberta}, are regarded as our prediction backbone, and we compare our framework \baby with the following four data augmentation methods:
\begin{itemize}
    \item \textbf{EDA} ~\cite{wei2019eda} is a token-level transformation method. It follows some patterns such as Synonym Replacement, Random Insertion, \etc For the ABSA task, we deliberately avoid corrupting the aspect words.
    \item \textbf{Back Translation (BT)}~\cite{sennrich-etal-2016-improving} is a sentence-level augmentation method, which translates a sentence to another language and translates it back to the original language. We follow \citet{fan2021beyond} that translates English to French and translates it back to English.
    \item \textbf{CBERT}~\cite{wu2019conditional} fully excavates the power of mask language model (MLM) objective to replace some tokens, so it is also a token-level replacement approach.
    \item \textbf{SCon}~\cite{liang2021enhancing} design aspect-invariant/-dependent data augmentation for ABSA and deploy a supervised contrastive learning objective. We reproduce it according to their released code and default best settings.
\end{itemize}


\subsection{More Empirical Details} \label{sec4.1}

We implement our experiment by following the released ABSA Pytorch\footnote{\url{https://github.com/songyouwei/ABSA-PyTorch}} repository  and our pre-trained models, such as \textit{bert-base-uncased}, \textit{roberta-base} and \textit{t5-base}, are from \textit{HuggingFace}\footnote{\url{https://huggingface.co/models}}. Meanwhile, we utilize a flexible toolkit \textit{OpenDelta}\footnote{\url{https://github.com/thunlp/OpenDelta}} to adapt various parameter-efficient methods to our T5 generator.

In the data augmentation stage, we leverage an \textit{Adafactor} optimizer. A T5 generator is fine-tuned for 100 epochs, and batch size is fixed to 16. As training a prediction model BERT or RoBERTa, we adapt an \textit{Adam} optimizer with a learning rate of $2 \times 10^{-5}$, and the dropout rate is set to 0.3. Additionally, we train the final prediction model for 15 epochs with the batch size as 16.

A single NVIDIA A100 is used to conduct our all experiments. The model fine-tuning and sentence generation stages will generally spend 2 - 3 hours, and it spends about 1 hour to train a prediction model.

\begin{figure*}[htbp]
  \centering
  \includegraphics[scale=0.7]{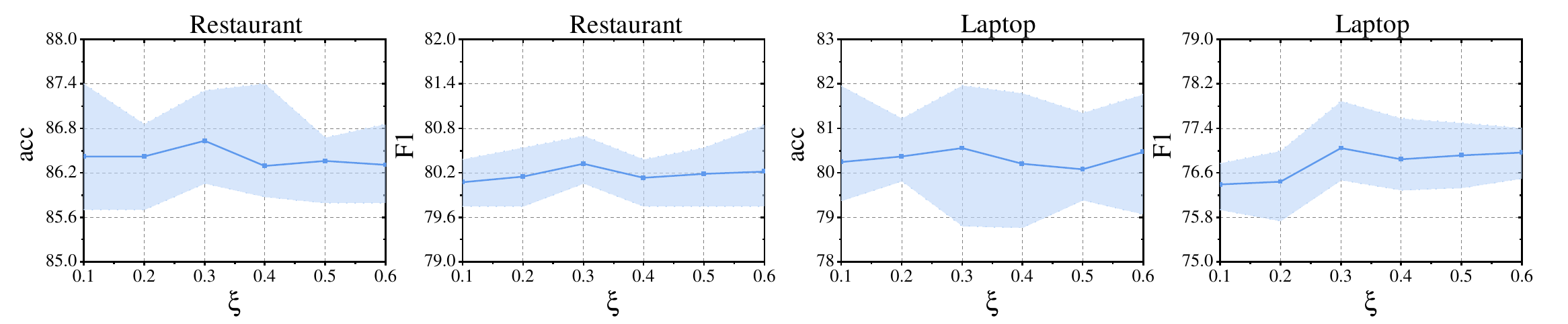}
  \caption{Sensitivity Analysis of the contrastive training margin $\xi$. The ribbon upper-side and below-side denote the maximum and minimum performance in 5 seeds respectively, and the centerline is the average value.}
  \label{xi}
\end{figure*}

\begin{figure*}[htbp]
  \centering
  \includegraphics[scale=0.7]{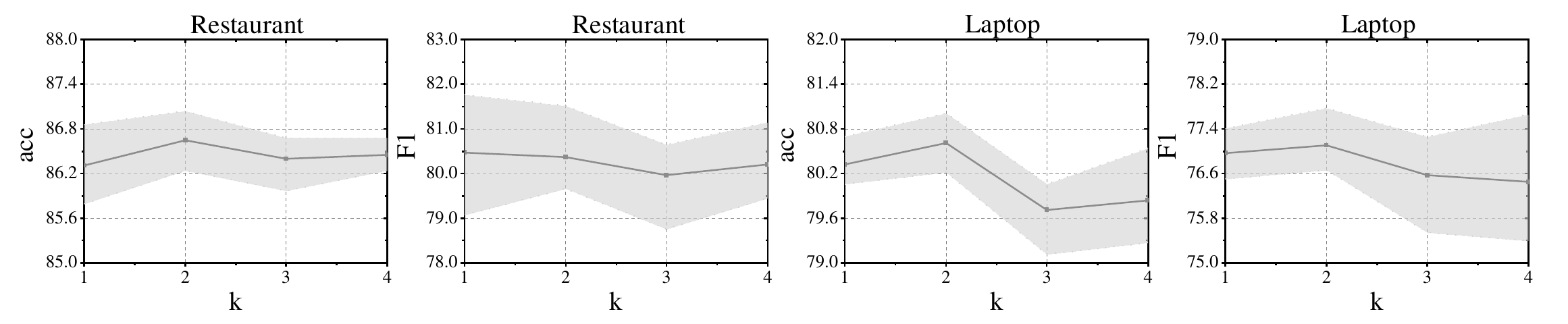}
  \caption{Sensitivity Analysis of the number of selected sentences $k$ in EMF.}
  \label{k}
\end{figure*}
\subsection{Main Result} \label{sec4.2}

Table~\ref{result} shows our main experimental results. To be fair, for each data augmentation method, we generate 1 synthetic sample per sentence. Overall, our \baby framework consistently and significantly improve the preformance of both BERT-based and RoBERTa-based models, e.g. the result of \textit{Restaurant} even outperforms the RoBERTa-\textit{base} baseline 2.22\% in term of F1. Moreover, we observe that the performance order of various data augmentation methods is \baby > CBERT > BT > EDA, which indicates: 
(1) EDA semantic-regardlessly removes or inserts tokens, which will inject more noise into final prediction models, thus leading to the poor performance.
(2) Comparing BT and CBERT, BT always achieves higher F1, and CBERT will get more advantageous accuracy. Since BT is a sentence-level method, it focuses on how to transform a source sentence into a form-different and semantic-reserved sentence, but neglects the useful aspect information. On the contrary, CBERT will only change some tokens, which is a semantic-reversed and aspect-reserved approach. 
Table~\ref{da} summarizes the characteristics of some augmentation methods.

Turn to compare two metrics, another finding is that \baby has more significant performance in term of Macro-F1. Proverbially, since Macro-F1 is a metric to measure the generalization of a model, the higher improvements on Macro-F1 prove that our framework can effectively improve the model's robustness in this view.
Concurrently, the effect of \baby is better on \textit{Restaurant} and \textit{Twitter}. The difference between these datasets is that \textit{Laptop} is a small dataset, indicating that our \baby is sensitive to the data scale and \baby may have better improvement if there is a larger-scale dataset.

\begin{table}[]
\centering
\renewcommand\arraystretch{1.1}
\small
  \caption{A comparison of various data augmentation methods. Semantic-R. and Aspect-R. denote semantic reserved methods and aspect reserved methods, and T.- / S.- discriminates token-level and sentence-level methods respectively.}
  \label{da}
  \setlength{\tabcolsep}{5pt}{
  \begin{tabular}{m{1.2cm}<{\centering}m{1.6cm}<{\centering}m{1.4cm}<{\centering}m{1.2cm}<{\centering}}
    \toprule
    Methods & Semantic-R. & Aspect-R. & T.- / S.- \\
    \midrule
    EDA & $\times$ & $\checkmark$ & T. \\
    BT & $\checkmark$ & $\times$ & S. \\
    CBERT & $\checkmark$ & $\checkmark$ & T. \\
    \textbf{\baby} & $\checkmark$ & $\checkmark$ & S. \\
    \bottomrule
  \end{tabular} }
\end{table}

\subsection{Ablation Study} \label{sec4.3}

\begin{table*}[]
\centering
\renewcommand\arraystretch{1.3}
\small
  \caption{Case study. The aspect words and corresponding polarities are noted as red color and blue color, respectively.}
  \label{case-distribution}
  \setlength{\tabcolsep}{5pt}{
  
  \begin{tabular}{m{3cm}<{\centering}m{1.3cm}<{\centering}m{1.3cm}<{\centering}m{1.3cm}<{\centering}m{1.3cm}<{\centering}m{1.3cm}<{\centering}m{1.3cm}<{\centering}}
    \toprule
    
    \rowcolor{lightgray} case & \multicolumn{6}{c}{\textit{the {\color{red} falafal} was rather over cooked and dried but the {\color{red} chicken} was fine.}} \\
    \midrule
    \multirow{2}{*}{Aspect \& Polarity} & \multicolumn{3}{c}{\textit{{\color{red} falafal}}} & \multicolumn{3}{c}{\textit{{\color{red} chicken}}} \\
    \cmidrule(lr){2-4} \cmidrule(lr){5-7}
    
    & \textit{Positive} & \textit{Neutral} & \textit{{\color{blue} Negative}} & \textit{{\color{blue} Positive}} & \textit{Neutral} & \textit{Negative} \\
    \hline
    w/o \baby & 0.098 & 0.109 & 0.793 & 0.643 & 0.047 & 0.310 \\
    with \baby & 0.071 & 0.092 & 0.837 & 0.698 & 0.076 & 0.226 \\

    \Xhline{0.75pt}
    \rowcolor{lightgray} case & \multicolumn{6}{c}{\textit{good {\color{red} food} but the {\color{red} service} was dreadful!}} \\
    \midrule
    \multirow{2}{*}{Aspect \& Polarity} & \multicolumn{3}{c}{\textit{{\color{red} food}}} & \multicolumn{3}{c}{\textit{{\color{red} service}}} \\
    \cmidrule(lr){2-4} \cmidrule(lr){5-7}
    
    & \textit{{\color{blue} Positive}} & \textit{Neutral} & \textit{Negative} & \textit{Positive} & \textit{Neutral} & \textit{{\color{blue} Negative}} \\
    \hline
    w/o \baby & 0.502 & 0.173 & 0.325 & 0.191 & 0.236 & 0.573 \\
    with \baby & 0.589 & 0.211 & 0.200 & 0.177 & 0.186 & 0.637 \\
    \bottomrule
  \end{tabular}
  }
\end{table*}
The core modules in our framework \baby are Data Augmentation (DA), Contrastive Learning (CL), and Entropy-Minimization Filter (EMF).
In this section, we perform ablation study to investigate the effect of these modules, and the results are listed in Table~\ref{ablation}. BERT-\textit{base} is regarded as the prediction model in this study, and \textit{w/o} denotes \textit{without} some modules.

Firstly, \textbf{w/o DA \& CL} removes our data augmentation framework, it is equivalent to a vanilla \textit{BERT-base}-only prediction model. As shown in Table~\ref{ablation}, w/o DA \& CL always presents poor performance, even decrease by about 1\% on \textit{Laptop}'s acc and F1 and \textit{Restaurant}'s F1. It proves our \baby framework imports more high-quality data items and is effective to boost the performance.
Then, \textbf{w/o CL} remains augmented sentences and revokes the contrastive training objective $\mathcal{R}_{CL}$. Overall, this objective improves the results by about 0.3\% on each dataset, conforming that contrastive learning can powerfully optimize the aspect-specific features. 
Finally, \textbf{w/o EMF} conducts both supervised training and contrastive training on all generated sentences. Intuitively, more samples can consolidate the predictive power of a classification model, but some emotionless and illogical sentences also carry plenty of noise. As we speculate, w/o EMF degrades the model's performance by 0.5\% - 1.0\%, the outcome on Restaurant's accuracy is even worse than the result without DA \& CL. More sensitivity studies about EMF will be shown in Section~\ref{sec4.4}.

Additionally, we also investigate the performance of generating sentences with \textbf{only AAC} and \textbf{only PAC} settings and report their results in Table~\ref{ablation}. We show that both them are beneficial, where only AAC slightly outperforms that of the only PAC, demonstrating that the multi-aspect data are more crucial than the multi-polarity data.  

\begin{table}[h]
\centering
\renewcommand\arraystretch{1.1}
\small
  \caption{Ablation study of \textbf{\baby}. The prediction model is BERT-\textit{base} in this study.}
  \label{ablation}
  \setlength{\tabcolsep}{5pt}{
  \begin{tabular}{m{1.8cm}<{\centering}m{1.4cm}<{\centering}m{1.4cm}<{\centering}m{1.4cm}<{\centering}}
    \toprule
    \multirow{2}{*}{Model} & \textit{Restaurant} & \textit{Laptop} & \textit{Twitter} \\
    \cmidrule{2-4}
    & \multicolumn{3}{c}{acc} \\
    \midrule
    \rowcolor{lightgray} \textbf{\baby} & \textbf{86.93} & \textbf{80.61} & \textbf{77.08} \\
    w/o DA \& CL & 86.31 & 79.66 & 76.50 \\
    w/o CL & 86.69 & 80.35 & 76.87 \\
    w/o EMF & 86.45 & 79.84 & 76.50 \\
    \hline
    only AAC & 86.63 & 80.10 & 77.16 \\
    only PAC & 86.44 & 79.94 & 77.19 \\
    \midrule
    \specialrule{0em}{0.5pt}{0.5pt}
    \midrule
    Model & \multicolumn{3}{c}{F1} \\
    \midrule
    \rowcolor{lightgray} \textbf{\baby} & \textbf{81.23} & \textbf{77.11} & \textbf{75.76} \\
    w/o DA \& CL & 80.22 & 76.11 & 75.23 \\
    w/o CL & 81.00 & 76.88 & 75.63 \\
    w/o EMF & 80.21 & 76.45 & 75.17 \\
    \hline
    only AAC & 80.61 & 76.69 & 75.93 \\
    only PAC & 80.24 & 76.17 & 75.91 \\
    \bottomrule
  \end{tabular} }
\end{table}

\subsection{Sensitivity Analysis} \label{sec4.4}
As mentioned above, we employ 4 hyper-parameters in our \baby framework, and $\xi$ and $k$ are more crucial. To investigate the influence of both hyper-parameters, we further conduct sensitivity analysis in this section. Additionally, we fix $\alpha$ and $\beta$ to 0.5 and 2.0, respectively. Limited by space, their sensitivity analysis will be described and analyzed in Appendix~\ref{appendixb}.

\paragraph{Contrastive Training Margin $\xi$}

Margin $\xi$ in Equation~\ref{Eq3} is an important hyper-parameter to control the model's distinctive ability to discriminate negative samples and positive samples. Under a little margin, the prediction model will indolently balance $d(\mathbf{h}_i, \mathbf{h}_i^p)$ and $d(\mathbf{h}_i, \mathbf{h}_i^n)$ to cause under-optimization, while a large margin makes a model converge hard.

To investigate the impact of the margin $\xi$, we implement experiments on \textit{Restaurant} and \textit{Laptop}, and range $\xi$ from 0.1 to 0.6. As shown in Figure~\ref{xi}, our \baby always gets the best performance when $\xi$ is 0.3, and as it expands or shrinks, the results consistently show a downward trend.

\paragraph{Number of Selected Sentences $k$ in EMF}

As the description in Section~\ref{sec4.3}, $k$ in EMF is a key hyper-parameter to decide to introduce more useful sentences or more noisy information, we select $k$ sentences for each source sentence with the entropy-minimization filter method.

We evaluate \baby on $k \in \{1,2,3,4\}$ and show the results in Figure~\ref{k}. The best performance invariably appears in $k=2$ or $k=1$. Meanwhile, the performance decreases significantly when the value of $k$ is 3, 4.

\subsection{Case Study}
\baby is a DA-based framework, and our purpose is to break the multi-aspect challenge. Therefore, we implement a case study to evaluate whether \baby can solve this problem, and more cases with different DA methods and cases with varying fine-tuning epochs will be shown in Appendix \ref{appendixc}.

We select two samples from the \textit{Restaurant} test set, and these two sentences all have two aspects with different sentiment polarities. We will compare the performance of our \baby on their polarity prediction distribution. To observe the Table~\ref{case-distribution}, we summarize that (1) \baby can consistently and significantly enhance the prediction model's robustness, such as the \textit{Positive} probability of the aspect \textit{food} improve about 0.077. Therefore, our framework can correct the polarity prediction when a sentence expresses an ambiguous polarity. (2) When a sentence is short, the interaction between two aspects with contradictory polarities will be more intimate. In this scenario, sentences will express more ambiguous sentiment and the effectiveness of our framework will be more significant.

\subsection{Computing Cost Analysis}

Our method will prolong each sentence, longer sentences will undoubtedly increase computing costs. Therefore, we attempt to compare the computing time of our framework and other augmentation methods. In the data preparing stage, \baby will fine-tune a generator and inference on the training set, it will spend 2 - 3 hours, and it is a once-for-all process. In the prediction training stage, we find that EDA, Back Translation and CBERT do not change the sentence length, their prediction time cost $0.98 \times \sim 1.05 \times$ (without any augmentation is $1 \times$). By comparison, \baby spends $1.14 \times$ time cost, which is completely acceptable.

\section{Conclusion}

In our work, we focus on solving the multi-aspect problem in the ABSA task. To address this challenge, we propose a data-centric training framework and design a novel \baby method to implement data augmentation. The proposed framework conduct both supervised training and contrastive training on augmented samples. The key idea of \baby is to generate some sentences that express the opposite sentiment polarity from the source sentences. Extensive experiments on widely-used benchmarks demonstrate that our framework can effectively improve the prediction model's robustness and predictive performance. In the future, it will be interesting to apply our \baby framework to improve other fine-grained natural language understanding tasks.

\section*{Acknowledgements}
We are grateful to the anonymous reviewers and the area chair for their insightful comments and suggestions. Bing Wang and Ximing Li were supported by the project from the National Natural Science Foundation of China (NSFC) (No.62006094).

\bibliography{C3DA}
\appendix
\begin{table*}[h]
\centering
\renewcommand\arraystretch{1.1}
\small
\caption{Case study. The upper table compares results of different data augmentation methods, the bold spans is changed part; The below table shows some generated sentences with \baby.}
  \label{case-tuning}
  \setlength{\tabcolsep}{5pt}{
  \begin{tabular}{m{2.1cm}<{\centering}m{12.8cm}<{\centering}}
    \toprule
    Method & Examples [dataset, polarity] \\
    \midrule
    \rowcolor{lightgray} source & \textit{not only was the {\color{red} food} outstanding, but the little {\color{red} 'perks'} were great.} {\color{blue} [Restaurant, Positive]} \\
    \textbf{No-tuning} & \textit{the {\color{red} kitchen} is very well equipped with all the {\color{red} equipment} you need to cook up a storm.} \\
    \textbf{Full-Tuning} & \textit{The {\color{red} food} is okay and the {\color{red} prices} here are mediocre.} \\
    \textbf{Prefix-tuning} & \textit{The {\color{red} kitchen} is a little small, but the {\color{red} food} is good.} \\
    \textbf{Prompt-tuning} & \textit{The {\color{red} kitchen} is a little dated, but the {\color{red} food} is good.} \\
    \textbf{LoRA} & \textit{The {\color{red} seats} are a bit cramped, but the {\color{red} food} was delicious.} \\
    
    
    \midrule
    \rowcolor{lightgray} source & \textit{it is easy to {\color{red} start up} and does not overheat as much as other laptops.} {\color{blue} [Laptop, Positive]} \\
    \textbf{No-tuning} & \textit{a {\color{red} laptop} that is easy to start up and does not overheat as much as other laptops.} \\
    \textbf{Full-Tuning} & \textit{I also purchased {\color{red} iWork} to go with it which has programs for word processing, spreadsheets, and presentations ( similar to Microsoft Office ).} \\
    \textbf{Prefix-tuning} & \textit{The {\color{red} battery} isn't very strong, but it is very light.} \\
    \textbf{Prompt-tuning} & \textit{but the {\color{red} screen} is so dark, and the {\color{red} service center} is a little numb.} \\
    \textbf{LoRA} & \textit{The {\color{red} software} is very easy to use, the {\color{red} start up} is very fast, the {\color{red} graphics} are fantastic.} \\
    \bottomrule
  \end{tabular} 
  }
\end{table*}

\section{Various Parameter-Efficient Methods Result} \label{AppendixA}

We adapt three parameter efficient methods for our T5 generator:
\begin{itemize}
    \item \textbf{Prefix-tuning} \cite{li2021prefix} fixes PLMs parameters, and inject an additional trainable prefix tokens for each transformer hidden layer. We set the number of prefix tokens to default 6.
    \item \textbf{Prompt-tuning} \cite{lester2021the} only prepends a prompt family to the input embedding layer. The length of the soft tokens is 100 in our experiments.
    \item \textbf{Low-Rank Adaptation (LoRA)} \cite{hu2021lora} trains a low-rank matrix to lightweight the PLMs. To be specific, we fix the hyper-parameter rank $r$ to 8 and cancel the dropout operation ($dropout = 0$). 
    \item \textbf{Full-tuning}  fine-tunes all parameters in the T5 backbone.
    \item \textbf{No-tuning} directly generates sentences with a pre-trained T5 model without fine-tuning.
\end{itemize}

\begin{figure}[h]
  \centering
  \includegraphics[scale=0.36]{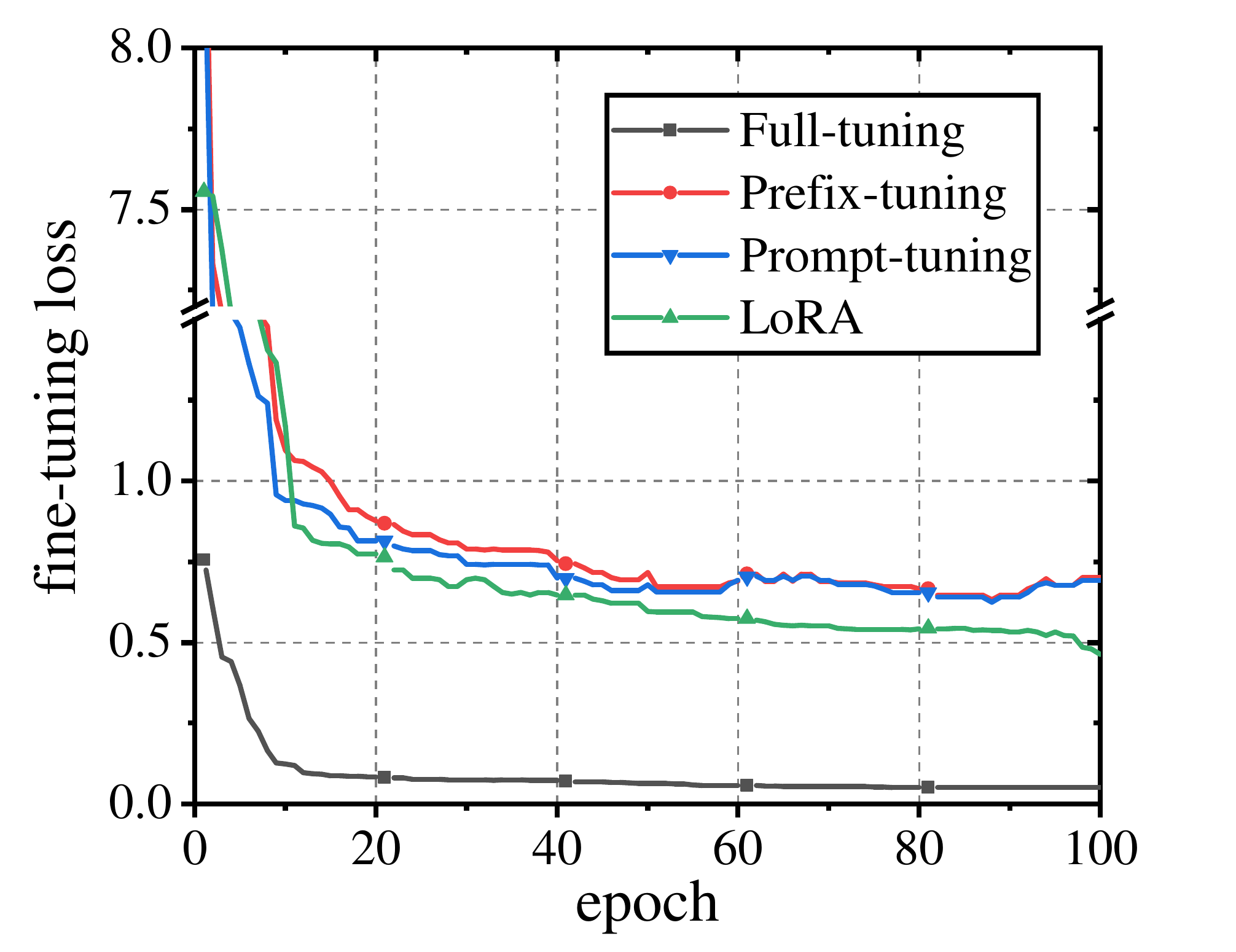}
  \caption{Convergence Analysis.}
  \label{convergence}
\end{figure}

\begin{table}[h]
\centering
\renewcommand\arraystretch{1.1}
\small
  \caption{Convergence loss. "-T." is short for "-Tuning".}
  \label{convergence-loss}
  \setlength{\tabcolsep}{5pt}{
  \begin{tabular}{m{0.7cm}<{\centering}m{1.3cm}<{\centering}m{1.3cm}<{\centering}m{1.3cm}<{\centering}m{1cm}<{\centering}}
    \toprule
    Method & Full-T. & Prefix-T. & Prompt-T. & LoRA \\
    \midrule
    Loss & 0.05102 & 0.70009 & 0.69182 & 0.46218 \\
    \bottomrule
    
  \end{tabular} 
  }
\end{table}

\begin{figure*}[h]
  \centering
  \includegraphics[scale=0.65]{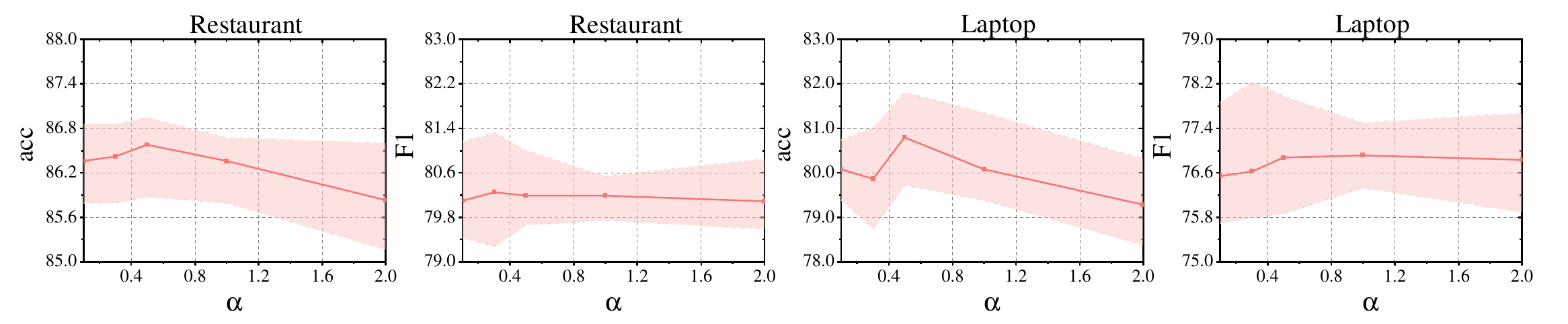}
  \caption{Sensitivity Analysis of the objective coefficient $\alpha$ in supervised classification training.}
  \label{alpha}
\end{figure*}
\begin{figure*}[h]
  \centering
  \includegraphics[scale=0.65]{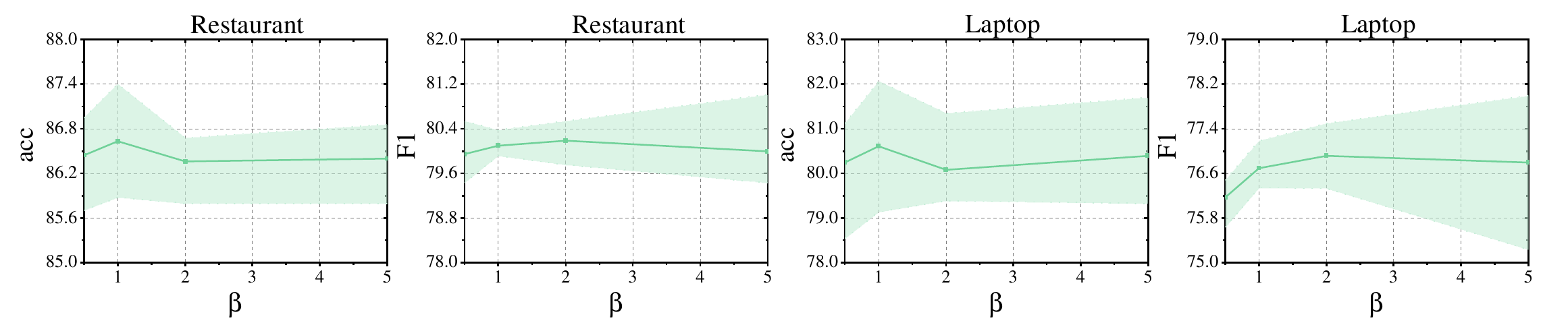}
  \caption{Sensitivity Analysis of the contrastive training objective coefficient $\beta$ in our overall objective.}
  \label{beta}
\end{figure*}

\subsection{Case Study}

More instances from different fine-tuning methods are listed in Table~\ref{case-tuning}, the domains and polarities are attached to the source sentences. Under our observation, we found some phenomena: (1) the sentences from Full-tuning are always longer, and Full-tuning habitually outputs the same sentences, which proves that Full-tuning will cause an over-fit problem by conjecture. (2) three parameter efficient methods consistently generate multi-aspect samples, however, it is doubtful whether more aspects in a sentence will promote the model's robustness more significantly. (3) most of the generated sentences can achieve our motivation (express an opposite polarity).

\subsection{Convergence Analysis}

We analyze the fine-tuning loss for those aforementioned fine-tuning methods (without No-tuning) with \texttt{T5ForConditionGeneration.loss} interface in \textit{Hugging Face} on \textit{Restaurant}. Figure~\ref{convergence} shows the loss curve, we adapt a percentile filter with 20 pts to conduct the curve smooth, and the losses of ultimate convergence are listed in Table~\ref{convergence-loss}. Obviously, Full-tuning will achieve a 
lower loss, because all parameters are tuned, and the scale of parameters is far greater than the number of data items, it is a precursor of an over-fitting problem. By comparing three fine-tuning methods, we prefer a method with lower convergence loss, Prefix-tuning and Prompt-tuning are indistinguishable, however, LoRA obtains an obvious superiority.

\section{More Sensitivity Analysis} \label{appendixb}

Objective coefficients $\alpha$ and $\beta$ are also controllable hyper-parameters, so we conduct sensitivity analysis on them. Our experimental results are shown in Figure~\ref{alpha} and Figure~\ref{beta}, the prediction model is consistently BERT-\textit{base} and $\alpha \in \{0.1, 0.3, 0.5, 1, 2\}, \beta \in \{0.5, 1, 2, 5, 10\}$. According to the figures, the best performance always when $\alpha = 0.3$ or $0.5$ and $\beta = 1$ or $2$.

\section{More Case Studies} \label{appendixc}

\paragraph{Different DA methods}

We investigate different data augmentation frameworks from the view of the model's prediction results in Section~\ref{sec4.2}. We will also list an example to compare these DA baselines and throw more generated sentences from our \baby in Table~\ref{case}. The example in the upper table is from \textit{Restaurant}, and we adapt LoRA to the T5 generator, and more examples of \baby below are from \textit{Restaurant} and \textit{Laptop}.

We summarize that \textbf{EDA} deletes \textit{"my"}, transforms \textit{"co-workers"} to its synonym \textit{"workers"} and randomly inserts a comma to an unreasonable position. This method adds some corruptions without any semantic relations, and it is possible to generate an illogical sample; Compared to EDA, \textbf{CBERT} is also a token-level method, but a semantic-reserved sentence is consistently created; \textbf{Back Translation} is a sentence-level method, and a correct sentence can be generated. However, it erases the aspect words for ABSA task. By comparison, our \textbf{\baby} not only reserves the origin semantic information but also introduces some high-quality in-domain corpus-aware information and adversarial signals. We also observe that most generated sentences from \baby exist two or more aspects, and this phenomenon is widespread in several datasets.

\begin{table*}[t]
\centering
\renewcommand\arraystretch{1.25}
\small
  \caption{Case study. The upper table compares results of different data augmentation methods, the blue spans is changed part; The below table shows some generated sentences with \baby.}
  \label{case}
  \setlength{\tabcolsep}{5pt}{
  
  \begin{tabular}{m{3cm}<{\centering}m{11cm}<{\centering}}
    \toprule
    DA methods & examples \\
    \midrule
    \rowcolor{lightgray} source & \textit{my co-workers had to wait almost an hour for delivery.} \\
    \textbf{EDA} & \textit{{\color{blue} workers} had to\textbf{,} wait an hour for delivery.} \\
    \textbf{B.T.} & \textit{{\color{blue} my colleague waited nearly an hour to deliver it.}} \\
    \textbf{CBERT} & \textit{my {\color{blue} mother} had to wait {\color{blue} half} an hour for delivery.} \\
    \textbf{\baby} & \textit{my co-workers had to wait almost an hour for delivery. {\color{blue} but the perks are great, and you get to enjoy all the perks.}} \\
    \bottomrule
  \end{tabular} 
  
  \begin{tabular}{m{7cm}<{\centering}m{7cm}<{\centering}}
    \toprule
    source sentences & \baby generated sentences \\
    \midrule
    \textit{if you love seafood, you would love this place!} & \textit{the {\color{red} sauce} was good, but the {\color{red} meats} were a little fatty.} \\
    \textit{we had a girls, night dinner here for restaurant week.} & \textit{try the {\color{red} samosas} at a reasonable price.} \\
    \textit{it is easy to start up and does not overheat as much as other laptops.} & \textit{but the {\color{red} screen} is so dark, and the {\color{red} service center} is a little numb.} \\
    \bottomrule
  \end{tabular} 
  }
\end{table*}

\paragraph{Cases with Varying Fine-tuning Epochs $\tau$}

We also explore the generation results with varying fine-tuning epochs $\tau$. As shown in Table~\ref{case-epoch}, given a source sentence ``\textit{but the staff was so horrible to us.}``, the T5 generator tends to generate an irrelative sentence when we have not yet fine-tuned it. In the middle of the fine-tuning stage, the generator is inclined to retell the source sentences or generate some irrational sentences. Finally, our generation model will eventually converge when $\tau=80$.

\begin{table}[h]
\centering
\renewcommand\arraystretch{1.2}
\small
  \caption{Case study. Given a source sentence ``\textit{but the staff was so horrible to us.}``, we can observe the trend of its augmented sentence with fine-tune epochs $\tau$.}
  \label{case-epoch}
  \setlength{\tabcolsep}{5pt}{
  \begin{tabular}{m{1.4cm}<{\centering}m{5.5cm}}
    \toprule
    epochs $\tau$ & sentence \\
    \hline
    \rowcolor{lightgray} source & \textbf{\textit{but the staff was so horrible to us.}} \\
    0 & \textit{i cried out for the suffragellah woman.} \\
    20 & \textit{the staff is so horrible to us.} \\
    40 & \textit{the as-is, the service isn't great.} \\
    60 & \textit{the food is very good.} \\
    80 & \textit{the food was great, the service was excellent.} \\
    \bottomrule
  \end{tabular} }
\end{table}

\section{Instance-level Loss Re-weight for Long-tail Aspects}

There is an another problem when we implement sentence generation, our generated sentences in one dataset often fit to few aspects, such as \textit{food} in \textit{Restaurant}. To investigate this issue, we show the aspect information for three public ABSA datasets in Table~\ref{frequency}, and we found that \textit{food} in \textit{Restaurant} actually maintain maximum frequency. Similarly, \textit{service}, \textit{battery} and \textit{screen} \etc report this long-tail distribution problem. And most of aspects in \textit{Twitter} is \textit{[UNK]}, it means that extensive aspect information will be unusable, this causes some models don't work, especially the attention-based models.

Similar to focal loss~\cite{lin2017focal}, we adapt an instance-level loss re-weight method to our fine-tuning stage. Inspired by \citet{jain2016extreme}, we multiple a multiplier to each instance's objective, the multiplier is shown in Equation~\eqref{Eq8}.

\begin{equation}
\label{Eq8}
    \Delta_{j} \ =\ \frac{1}{1 + C e^{-A\log(M_j + B)}},
\end{equation}

\begin{equation}
\label{Eq9}
    C \ =\ (\log M - 1)(B + 1)^A,
\end{equation}
where $M_j$, $M$ are the number of data points annotated with aspect $j$, and aspect items, respectively.

\begin{table}[h]
\centering
\renewcommand\arraystretch{1.2}
\small
  \caption{The aspect words and its frequency statistic of three public datasets. We only show the top-10 aspect words (with the T5 tokenizer) and hide some special tokens such as '\textit{a}', '\textit{i}', \etc}
  \label{frequency}
  \setlength{\tabcolsep}{5pt}{
  \begin{tabular}{m{1.6cm}<{\centering}m{5.3cm}<{\centering}}
    \toprule
    dataset & aspect words: frequency \\
    \midrule
    \textit{Restaurant} & \textit{food}: 419, \texttt{[UNK]}: 404, \textit{service}: 204, \textit{wait}: 94, \textit{menu}: 80, \textit{dinner}: 75, \textit{wine}: 72, \textit{staff}: 68
    , \textit{pizza}: 65, \textit{place}: 60 \\
    \hline
    
    \textit{Laptop} & \texttt{[UNK]}:267, \textit{battery}: 97, \textit{screen}: 81, \textit{use}: 58, \textit{life}: 56, \textit{windows}: 55, \textit{price}: 55, \textit{software}: 54
    , \textit{keyboard}: 54, \textit{drive}: 53 \\
    \hline
    
    \textit{Twitter} & \texttt{[UNK]}: 5328, \textit{spear}: 894, \textit{bri}: 887, \textit{ney}: 887, \textit{lo}: 394, \textit{han}: 393, \textit{lind}: 388, \textit{say}: 388
    , \textit{in}: 370, \textit{ry}: 360 \\
    \bottomrule
  \end{tabular} }
\end{table}

\end{document}